\documentclass{bmvc2k}
\usepackage{booktabs} 
\usepackage{multirow}
\usepackage{wrapfig}

\title{Cross-Task Generalization Between Understanding and Generation in Unified Vision-Language Models: A Controlled Study}

\addauthor{Jihai Zhang\textsuperscript{*}}{jihaizhang@link.cuhk.edu.hk}{1}
\addauthor{Tianle Li}{tianleli@link.cuhk.edu.hk}{1}
\addauthor{Linjie Li}{}{2}
\addauthor{Zhengyuan Yang}{}{2}
\addauthor{Yu Cheng}{https://ych133.github.io}{1}

\addinstitution{
 The Chinese University of Hong Kong\\ Hong Kong, China
}
\addinstitution{
 Microsoft \\ Seattle, Washington, USA
}

\runninghead{Zhang et al.}{Understanding-Generation Transfer in Unified VLMs}

\begin{document}
\maketitle

\begingroup
  \makeatletter
  \renewcommand{\@makefntext}[1]{\noindent#1}
  \makeatother
  \footnotetext{\textsuperscript{*}\ This work was completed during Jihai
  Zhang's internship at Microsoft.}
\endgroup

\begin{abstract}
Unified vision-language models (VLMs) aim to support both visual understanding and generation within a single framework, but it remains unclear when mixed training benefits both capabilities and when it introduces conflicts.
This paper presents a controlled empirical study of cross-task generalization between understanding and generation in unified VLMs.
We construct two controllable image-text benchmarks, SmartWatch and modified CelebA, with paired VQA, captioning, and text-to-image generation tasks, and evaluate multiple LLM-based unified architectures built from SigLIP and VQ-VAE visual spaces.
Our experiments show that mixed understanding-generation training can improve both tasks over task-specific training, but the benefit depends strongly on the relation between vision input and output spaces.
Unified models with better aligned visual spaces exhibit stronger cross-task transfer, while reversible affine distortions of the input visual space substantially weaken this effect and can turn mutual benefits into conflicts.
We further find that increasing data from one task can initially improve the other, but excessive imbalance between understanding and generation data may degrade the complementary task.
By controlling attribute frequencies, we show that generation supervision can help recover underrepresented visual concepts for understanding.
Adapter analyses suggest that this transfer is not primarily caused by richer visual adapter features, but by the base language model learning relationships that generalize across aligned visual spaces.
A real-case experiment on LLaVA provides additional evidence that mixed understanding-generation training can benefit visual understanding beyond controlled benchmarks.  Our source code is available at \href{https://github.com/MajorDavidZhang/Generalization_unified_VLM}{github.com/MajorDavidZhang/Generalization\_unified\_VLM}.
\end{abstract}

\section{Introduction}
\label{sec:intro}

In recent years, Vision-Language Models (VLMs) have emerged as a central paradigm in artificial intelligence. These models are often categorized into two broad families: \textit{understanding VLMs}~\citep{chen2024expanding,liu2023visual,wang2024qwen2}, which focus on tasks such as visual question answering (VQA) and image captioning; and \textit{generation VLMs}~\citep{betker2023improving,podell2023sdxl,tian2024visual}, which are designed for image generation and editing.
Although specialized models have achieved strong performance in their respective domains, recent research has increasingly explored \textit{unified VLMs}~\citep{chen2025janus,xie2024show,zhou2024transfusion,wang2024emu3,team2024chameleon}, which aim to support understanding and generation within a single model.

The motivation behind unified VLMs is the hypothesis that shared parameters and mixed training across understanding and generation may enable cross-task transfer. As Richard Feynman famously stated, ``What I cannot create, I do not understand.'' In the context of VLMs, this intuition suggests that concepts learned from understanding tasks may help generation, while generation supervision may also encourage the model to learn relationships useful for understanding.
For example, spatial or attribute concepts learned from captions and VQA may help a model generate images that better follow text instructions.
Conversely, learning to generate images from text may require the model to associate textual concepts with visual representations, which could improve its ability to answer related visual questions.

\begin{wrapfigure}{r}{0.5\textwidth}  
  \centering
  \includegraphics[width=0.49\textwidth]{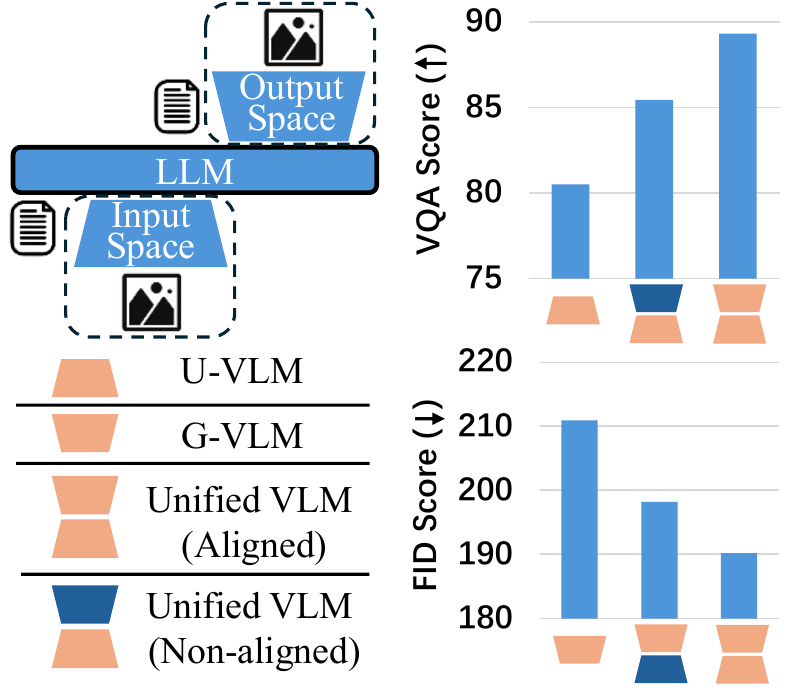}  
  \caption{Mixed understanding-generation training can improve task performance under controlled settings, especially when vision input and output spaces are aligned. Results from Section~\ref{sec:exp}.}
  \label{fig:intro}
\end{wrapfigure}

Despite the growing interest in unified VLMs, most existing works focus on architectural design or training recipes, while the interaction between understanding and generation remains less clearly characterized. Existing unified VLMs do not always show clear improvements over specialized models on every task~\citep{chen2025janus,team2024chameleon}, and prior analyses report mixed conclusions about whether understanding and generation help or interfere with each other~\citep{tong2024metamorph,chen2025janus,wu2024liquid}.
This raises a more specific question: under what conditions does mixed understanding-generation training produce cross-task generalization, and when can it introduce conflicts?
To study this question, we conduct a controlled empirical study of cross-task generalization between understanding and generation in unified VLMs. We construct two controllable image-text benchmarks, SmartWatch and modified CelebA, that provide paired understanding data (VQA and captioning) and generation data (text-to-image generation). We then evaluate multiple LLM-based unified VLM architectures built from SigLIP and VQ-VAE visual spaces. Our key findings are as follows:

\emph{First, mixed training can produce cross-task benefits, but the effect is not unconditional.}
Compared with task-specific baselines trained on matched data, most mixed understanding-generation models improve the target task, as illustrated in Figure~\ref{fig:intro}. By varying the amount of understanding and generation data, we further observe that adding data from one task can initially improve the other task, but excessive imbalance may lead to trade-offs rather than monotonic gains.

\emph{Second, alignment between vision input and output spaces is a key factor for cross-task generalization.}
Unified VLMs with better aligned vision input and output spaces tend to exhibit stronger transfer. To test this factor, we introduce reversible affine distortions to the vision input space. These distortions have little effect on task-specific models but substantially weaken mixed-training benefits, indicating that cross-task transfer depends on the relation between the two visual spaces.

\emph{Third, generation supervision can transfer visual concepts to understanding.}
Using the controllability of our datasets, we create settings where selected attributes are rare in understanding data but remain present in generation data. In these cases, mixed models recover the underrepresented concepts much better than understanding-only baselines. Further analysis suggests that the transfer is not primarily caused by richer modality-adapter features, but by the base language model learning relationships that generalize across aligned visual spaces.

Overall, our results provide controlled evidence that understanding and generation can benefit each other in unified VLMs, while also showing that such benefits depend on architecture, visual-space alignment, and data balance. We hope this study provides useful guidance for designing unified VLMs and for evaluating cross-task interactions in future work.

\section{Related Work}
\label{sec:related_work}

\noindent{\textbf{Unified Vision-Language Models.}}
Recent efforts have aimed to build unified VLMs that support visual understanding and generation within one modeling framework. A major line of work extends autoregressive language modeling to multimodal sequences, using visual tokenizers to place images and text into a common next-token prediction interface~\citep{wu2024vila,wang2024emu3,liu2024world,team2024chameleon,team2026longcat}. This formulation is attractive because it exposes understanding, generation, and interleaved multimodal reasoning to the same backbone, creating the possibility that supervision from one task can benefit the other.

A central challenge is that unification must make the visual representations used for understanding compatible with those produced for generation. Existing models address this with different task interfaces, including diffusion-based decoders or hybrid autoregressive-diffusion objectives~\citep{dong2023dreamllm,tian2024mm,zhou2024transfusion,xie2024show,ma2024janusflow}, shared visual-textual representations~\citep{wu2024liquid}, and partially separated pathways for reducing interference~\citep{li2024dual,chen2025janus}. Recent work on architecture decoupling further shows that stronger decoupling can improve unified models by making their cross-modal interaction patterns closer to task-specific understanding and generation models~\citep{zheng2025architecture}. These studies suggest that effective unification depends not only on shared parameters, but also on how the two tasks are made compatible. Our work studies this compatibility from a controlled representation-space perspective by varying the relation between the vision input and output spaces and measuring the resulting cross-task transfer.\\

\noindent{\textbf{Generalization Across Generation and Understanding.}}
Although unified VLMs are motivated by potential mutual enhancement between understanding and generation, existing evidence shows that this benefit is conditional. \citet{tong2024metamorph} report consistent mutual gains in a unified VLM that uses SigLIP~\citep{zhai2023sigmoid} for understanding and generates SigLIP tokens that condition an image decoder, suggesting that aligned semantic visual spaces can support bidirectional transfer. In contrast, \citet{wu2024liquid} observe impairment when VQ-VAE~\citep{van2017neural} is used for both visual input and output tokenization, with the impairment reduced only as the model size increases. Janus-Pro further shows that decoupling the visual pathways can alleviate conflicts, yet the interaction between understanding and generation remains mixed across tasks~\citep{chen2025janus}. These results indicate that cross-task generalization is not an automatic consequence of unified training; it depends on visual representation, architecture, scale, and data mixture.

Recent benchmarks have made this question more explicit. RealUnify directly evaluates whether unified models truly benefit from unification by measuring bidirectional synergy between understanding and generation~\citep{shi2025realunify}. MME-Unify broadens evaluation to understanding, generation, and mixed-modality generation within one benchmark~\citep{xie2025mme}, while UEval focuses on unified multimodal generation tasks that require models to produce both text and images~\citep{li2026ueval}. These efforts reflect a shift from asking whether a model can perform both tasks to asking whether the unified formulation actually improves the interaction between them.

Recent training methods also exploit cross-task transfer more directly. UniMRG uses auxiliary multi-representation generation targets to improve visual understanding, indicating that generation supervision can strengthen perception when it captures useful visual structure~\citep{su2026generation}. Conversely, UNO uses understanding-oriented supervision to steer visual generation, showing that semantic and structural signals from understanding can improve generation and editing~\citep{liu2026steering}. These works demonstrate the practical value of transfer in both directions, but they are typically studied within specific large systems and training recipes. Taken together, recent evaluation and training studies suggest that cross-task synergy is becoming a central criterion for unified VLMs, yet the factors that make the synergy appear or disappear remain difficult to separate in full-scale systems. Our work complements them with a controlled empirical study that fixes the task format and systematically varies visual input/output alignment and data balance, allowing us to analyze when mixed understanding-generation training helps and when it introduces conflict.

\section{Preliminaries}
\label{sec:prelim}

\subsection{Datasets}
To isolate the factors that affect cross-task generalization, we use two controllable image-text benchmarks: a synthetic SmartWatch dataset and a modified CelebA~\citep{liu2015faceattributes} dataset. The goal of these datasets is not to benchmark state-of-the-art generation or understanding performance, but to create paired understanding and generation tasks with shared semantic attributes, controlled data distributions, and low experimental cost. Unless otherwise specified, we use 60K VQA samples, 60K captioning samples, and 60K text-to-image generation samples as the default training set.

The two datasets play complementary roles in our analysis. SmartWatch gives us complete control over both continuous attributes (time and battery) and discrete attributes (weather and positions), making it suitable for controlled distribution shifts and rare-attribute interventions. CelebA uses real images while retaining attribute-level supervision, allowing us to test whether the same trends appear beyond rendered synthetic images. Both datasets share the same paired task format: each semantic attribute can appear in VQA, captioning, and text-to-image generation data. This shared format lets us vary the task mixture while keeping the underlying visual concepts fixed.\\

\begin{figure}[!ht]
    \centering
    \subfigure[Samples from the SmartWatch dataset showing time 02:07:00, cloudy weather displayed at the top-right, and 51\% battery displayed at the bottom-left.]{
        \label{fig:sw-1}
        \includegraphics[width=0.48\textwidth]{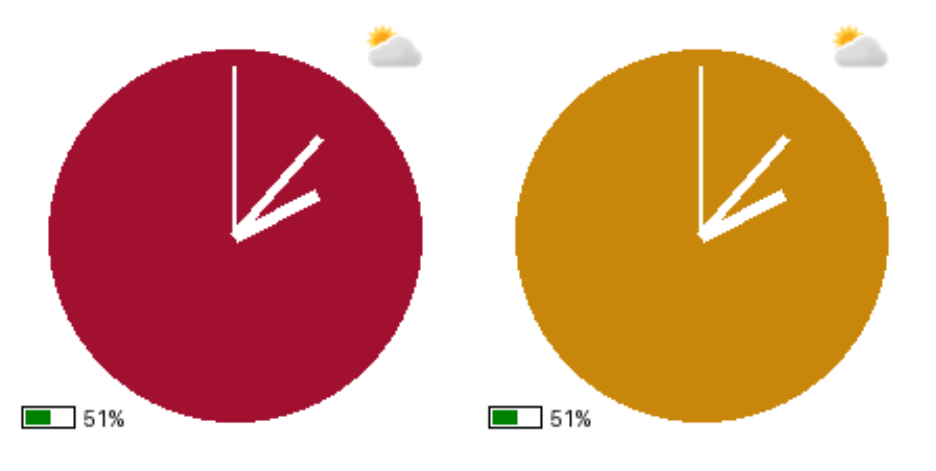}
    }
    \hfill
    \subfigure[Samples from the CelebA dataset showing person who does not have black hair, has no beard, is a female, is not wearing a hat, is not wearing a necklace.]{
        \label{fig:sw-2}
        \includegraphics[width=0.48\textwidth]{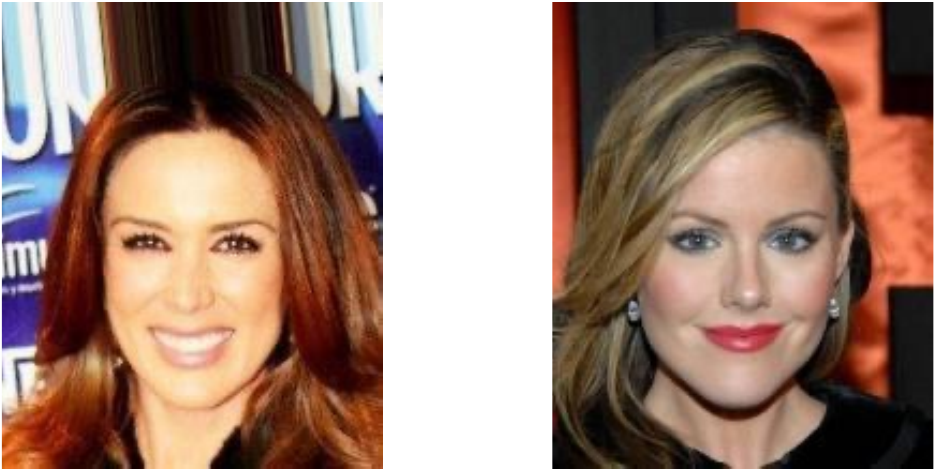}
    }
    \caption{Samples from SmartWatch and modified CelebA with corresponding ground-truth attributes. SmartWatch provides fully controlled rendered images, while CelebA provides real images with controlled attribute descriptions.}
    \label{fig:sw}
\end{figure}

\noindent{\textbf{SmartWatch.}}
SmartWatch provides a fully controllable setting for both image understanding and generation tasks, as illustrated in Figure~\ref{fig:sw}.
Each image in the dataset is controlled by six distinct attributes: time, weather, weather position, battery level, battery position, and watch face color. Specifically:
Time consists of hour, minute, and second values ranging from 0--12, 0--60, and 0--60, respectively.
Weather can take on three states: cloudy, rainy, or sunny.
Battery level ranges from 0 to 100.
Both weather position and battery position can be top-left, top-right, bottom-left, or bottom-right.
To generate each data sample, we first define the ground truth for the six attributes. A rule-based generator then creates the corresponding image based on these ground truth attributes, ensuring accurate pairing between image and text data. 
The watch face color is randomly sampled and not included in any text data, allowing a single text description to correspond to multiple visual appearances. This design separates task-relevant attributes from nuisance visual variation, which is useful for studying whether knowledge transfers across understanding and generation. For understanding tasks, we generate VQA and captioning data using diverse QA templates. Each VQA question focuses on one attribute, with equal appearance probabilities for the five task-relevant attributes. For captioning data, each caption includes 1--5 attributes, with time always present and the other four appearing independently with probability 0.5. For generation tasks, we create text-to-image instructions following the captioning templates.\\

\noindent{\textbf{CelebA.}}
CelebA complements SmartWatch with real face images while retaining attribute-level control. We select seven clearly defined attributes in the original CelebA: black hair, eyeglasses, no beard, male, wearing hat, wearing necklace, and wearing necktie. To generate one image-text pair, we first sample the target attributes, generate text using predefined templates, and then filter and randomly sample a matching image. Training and testing images come from non-overlapping image pools. We generate VQA, captioning, and text-to-image data using the same procedure as SmartWatch, which lets us compare controlled synthetic images with real-image attribute data under a matched task format.

\subsection{Evaluation}
We evaluate understanding tasks in the VQA format and report VQA accuracy. For SmartWatch, we compute scores over four key attributes: time, weather, position, and battery. Weather and position are evaluated by exact matching. Time and battery have larger value ranges, so exact matching alone would not reflect partial progress during training. We therefore compute time accuracy as $1 - \frac{\text{error}_h}{3} - \frac{\text{error}_m}{3} - \frac{\text{error}_s}{3}$, where $\text{error}_h = \min(|h-\hat h|, 12-|h-\hat h|)/6$, with analogous minute/second metrics. And battery accuracy as $1 - \frac{\text{error}_b}{100}$. For CelebA, attribute questions are evaluated by matching the predicted answer to the ground-truth attribute label. For generation tasks, we compute the FID score~\citep{heusel2017gans} between the ground-truth images and the generated images.

\subsection{Unified VLMs}
In this paper, we primarily focus on LLM-based Unified Vision-Language Models (VLMs). For image understanding, a pre-trained image encoder first encodes the input image. Subsequently, a dedicated understanding vision adapter projects the encoded representations from the encoder's hidden space to the LLM's hidden space, transforming them into input vision tokens. These vision tokens are then combined with text tokens and fed into the LLM. The pre-trained image encoder can be either a VQ-VAE encoder\citep{van2017neural} or a SigLIP vision encoder \citep{zhai2023sigmoid}.
For image generation, the LLM generates a special symbol, ``<image>'', at the beginning of the generation process. Upon encountering this symbol, an image generation head is activated instead of the language modeling head. This image generation head projects the LLM's output hidden states into vision tokens. After generating the first vision token, a generation vision adapter maps these tokens back to the LLM's input hidden space to enable autoregressive generation. The parameters of the generation vision adapter can optionally be shared with those of the understanding vision adapter, depending on whether the input and generated vision tokens reside in the same latent space. The generated vision tokens may correspond to SigLIP vision embeddings or VQ-VAE tokens.

\subsection{Experiment Settings}
We evaluate various combinations of VQ-VAE and SigLIP within unified VLMs, allowing us to compare models with aligned and misaligned vision input/output spaces:

\textbf{SigLIP-VQ.} SigLIP serves as the vision encoder for the VLM, and the model generates VQ token IDs that are decoded into real images by the VQ-VAE decoder, following the input/output-space pattern used by Janus~\cite{chen2025janus}.
    
\textbf{VQ-VQ.} The VQ-VAE encoder acts as the vision encoder for the VLM, generating VQ token IDs that are decoded into real images by the VQ-VAE decoder, resembling Liquid~\cite{wu2024liquid}.
    
\textbf{SigLIP-SigLIP.} SigLIP is used as the vision encoder for the VLM, and the model generates SigLIP embeddings, following the aligned-space setting studied by MetaMorph~\cite{tong2024metamorph}.
    
\textbf{VQ-SigLIP.} The VQ-VAE encoder serves as the vision encoder for the VLM, generating SigLIP embeddings. While this configuration is not practical for real-world applications, it serves as a valuable baseline for comparison in our experiments.

For VQ-VAE, we use \texttt{vq\_ds16\_t2i} from \citet{sun2024autoregressive}, with a resolution of $256 \times 256$. For SigLIP, we use \texttt{SigLIP-base-patch16-224} from \citet{zhai2023sigmoid}, with a resolution of $224 \times 224$. 
In unified VLMs with aligned input and output vision spaces, we share the parameters of the understanding vision adapter and generation vision adapter by default to maintain alignment. For the base LLM in all unified VLMs, we use \texttt{Vicuna-7B-v1.5}~\citep{peng2023instruction} for fair comparison. We adopt a one-stage training approach to jointly update the vision adapters, image generation head, and LLMs. By default, we fine-tune the base LLM using Low-Rank Adaptation (LoRA).

For experiments on SmartWatch and CelebA, training is performed for one epoch with a global batch size of 393 on 3 Nvidia H100 80GB GPUs. We use LoRA with rank 128, $\alpha=256$, and learning rate $2e^{-4}$ for the base LLM. The learning rates for the understanding vision adapter, generation vision adapter, and image generation head are set to $1e^{-4}$. For each dataset, we independently generate 1.5K test samples, including 1K VQA samples and 500 text-to-image generation samples. For models that generate CLIP or SigLIP embeddings, we use cosine similarity loss for the generation part. For models that generate VQ-VAE token IDs, we use cross-entropy loss. For SigLIP-VQ, we set the generation loss weight to 0.2 to account for the scale difference between VQ-token cross-entropy and language-token cross-entropy.

\section{Generalization Across Understanding and Generation}
\label{sec:exp}
\begin{figure}[!ht]
    \centering
    \subfigure[Image understanding performance of SigLIP-as-vision-encoder unified VLM.]{
        \label{fig:u-SigLIP}
        \includegraphics[width=0.30\textwidth]{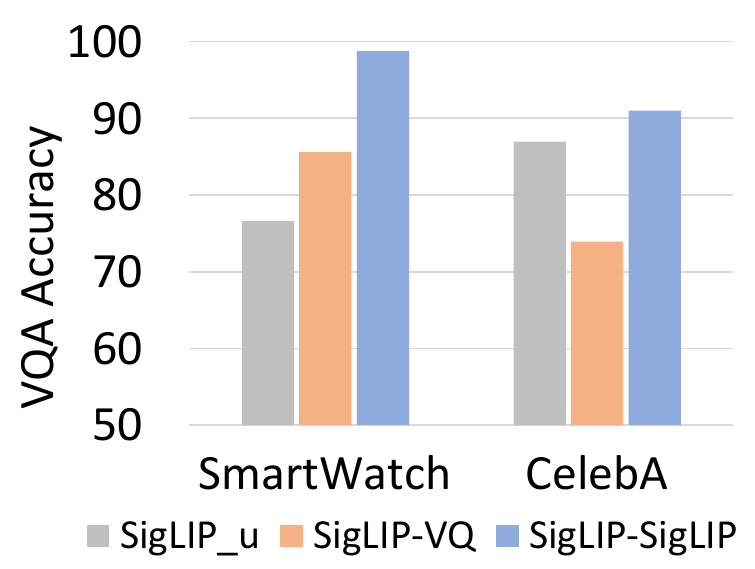}
    }
    \hfill
    \subfigure[Image understanding performance accuracy of VQVAE-as-vision-encoder unified VLMs.]{
        \label{fig:u-vq}
        \includegraphics[width=0.30\textwidth]{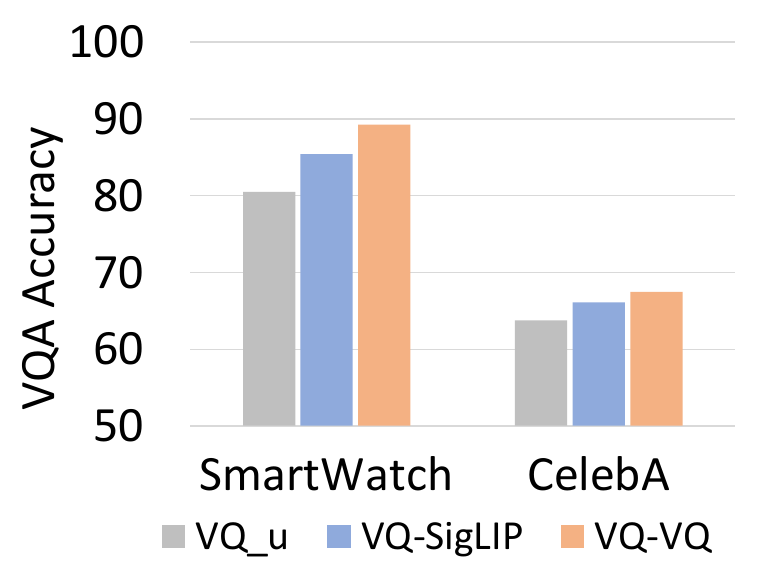}
    }
    \hfill
    \subfigure[Image generation performance of VQVAE-as-vision-decoder unified VLMs.]{
        \label{fig:g-vq}
        \includegraphics[width=0.30\textwidth]{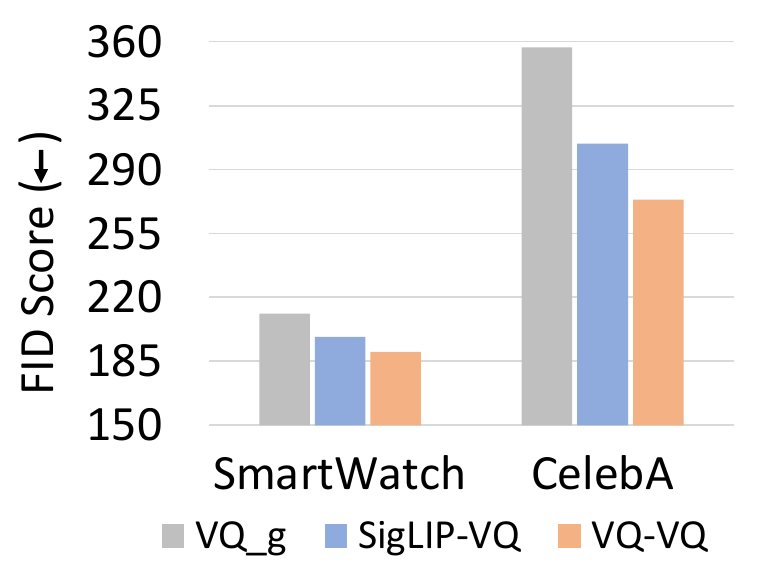}
    }
    \caption{Image understanding and generation performance of VLMs during training. ``\_g'' denotes generation-only training, and ``\_u'' denotes understanding-only training. Mixed training often improves over task-specific baselines, especially when the vision input and output spaces are aligned (SigLIP-SigLIP and VQ-VQ).}
    \label{fig:eval-1}
\end{figure}

We first train the four unified VLM variants using a mixture of understanding and generation data, and compare them with task-specific models trained only on understanding or generation data. The evaluation results are shown in Figure~\ref{fig:eval-1}.
Most mixed-training models outperform their task-specific counterparts on the corresponding target tasks. For understanding, SigLIP-SigLIP surpasses SigLIP\_u, SigLIP-VQ surpasses SigLIP\_u on SmartWatch, and VQ-VQ and VQ-SigLIP outperform VQ\_u. For generation, both SigLIP-VQ and VQ-VQ improve over VQ\_g. These results provide controlled evidence that cross-task generalization can occur between understanding and generation, but the strength of the effect varies across architectures and datasets. 

As an additional sanity check on whether the understanding-only
baselines on SmartWatch are simply underfit, we evaluate the larger Qwen3-VL-235B-A22B-Instruct on the SmartWatch benchmark. It achieves 100.0/86.7/72.3/100.0/89.6 on weather/position/time/battery/total, respectively. Its remaining errors include minor position hallucination and confusion involving the second hand, suggesting that the time attribute remains non-trivial even for a much larger VLM. This comparison
supports that the mixed-training gains are not simply an artifact of underfitting in the understanding-only baselines.

\begin{table*}[!htbp]

\caption{Effect of the affine transformation on model performance. SW and CA denote the SmartWatch and CelebA datasets, respectively. Values in parentheses indicate the performance change after applying the transformation. The reversible transformation has little effect on task-specific baselines but weakens mixed-training models, indicating that cross-task transfer depends on the relation between vision input and output spaces.}
\centering
\footnotesize
\setlength{\tabcolsep}{3pt}
\begin{tabular}{c|l|cc|cc|cc}
\toprule
& & \multicolumn{2}{c}{\textbf{SigLIP\_u}} & \multicolumn{2}{|c}{\textbf{VQ\_u}} & \multicolumn{2}{|c}{\textbf{VQ\_g}} \\ 
 \textbf{Model Type} & \textbf{Transform} & \textbf{SW VQA $\uparrow$} & \textbf{CA VQA $\uparrow$} & \textbf{SW VQA $\uparrow$} & \textbf{CA VQA $\uparrow$} & \textbf{SW FID $\downarrow$} & \textbf{CA FID $\downarrow$} \\ 

\midrule
\multirow{2}{*}{Non-Unified} & w\textbackslash o-Affine & 76.6 & 87.0     & 80.5 & 63.8 & 210.9 & 357.0    \\ 
& w-Affine                   & 78.2(+1.6) & 87.3(+0.3)   & 78.9(-1.6) & 63.3(-0.5) & 210.9(-0.0) & 357.0(-0.0)    \\ 
\midrule
& & \multicolumn{2}{c|}{\textbf{SigLIP-SigLIP}} & \multicolumn{4}{c}{\textbf{VQ-VQ}} \\
\textbf{Model Type} & \textbf{Transform} & \textbf{SW VQA $\uparrow$} & \textbf{CA VQA $\uparrow$} & \textbf{SW VQA $\uparrow$} & \multicolumn{1}{c}{\textbf{CA VQA $\uparrow$}} & \multicolumn{1}{c}{\textbf{SW FID $\downarrow$}} & \textbf{CA FID $\downarrow$} \\ 
\midrule
\multirow{2}{*}{Unified} & w\textbackslash o-Affine   & 98.7  & 91.0      & 89.3 & \multicolumn{1}{c}{66.1} & \multicolumn{1}{c}{190.2} & 273.5  \\ 
& w-Affine                     & 91.3(-7.5)& 85.3(-5.8)   & 85.1(-4.2) & \multicolumn{1}{c}{64.5(-1.6)} & \multicolumn{1}{c}{195.8(+5.6)}      & 368.4(+94.9) \\
\bottomrule
\end{tabular}
\label{tab:affine}
\end{table*}

From Figure~\ref{fig:eval-1}, we observe that models with aligned vision input and output spaces tend to show stronger mixed-training benefits. Specifically, SigLIP-SigLIP outperforms SigLIP-VQ in understanding tasks, VQ-VQ surpasses VQ-SigLIP in understanding tasks, and VQ-VQ outperforms SigLIP-VQ in generation tasks. This motivates us to examine whether cross-task generalization is influenced by the relation between the vision input and output spaces.

To test this factor, we introduce a random affine transformation immediately after the understanding vision adapter. This transformation distorts the input visual space relative to the output visual space while remaining reversible, reducing the risk that performance changes are simply caused by information loss. The results are shown in Table~\ref{tab:affine}.
The affine transformation has little effect on task-specific understanding models, suggesting that the transformed input representation remains usable for understanding. In contrast, mixed-training models perform worse after the transformation, especially for aligned-space models such as SigLIP-SigLIP and VQ-VQ. The degradation appears in both understanding and generation metrics, and in some cases makes mixed-training models worse than their task-specific counterparts. These results suggest that understanding can benefit from generation only when the model can relate the visual representations used for input and output. When the two spaces are misaligned, the base LLM may struggle to reuse relationships learned from one task in the other, reducing transfer or even causing task interference.

This intervention is useful because it separates two possible explanations for performance changes. If the affine transformation simply destroyed useful visual information, task-specific understanding models would also degrade substantially. Instead, their performance remains nearly unchanged, which indicates that the transformed representation still preserves the information needed for the understanding task. The degradation is concentrated in mixed-training models, where the model must use relationships learned from both input-side and output-side visual representations. This pattern suggests that the main issue is not whether the input representation is informative in isolation, but whether the LLM can align it with the visual space used for generation. The result therefore supports that visual-space alignment is a mechanism for cross-task transfer, not merely an implementation detail.
\begin{figure}[!ht]
    \centering
    \subfigure[Performance of understanding and generation as the amount of understanding data increases, with generation data fixed at 60K.]{
        \label{fig:more_un}
        \includegraphics[width=0.47\textwidth]{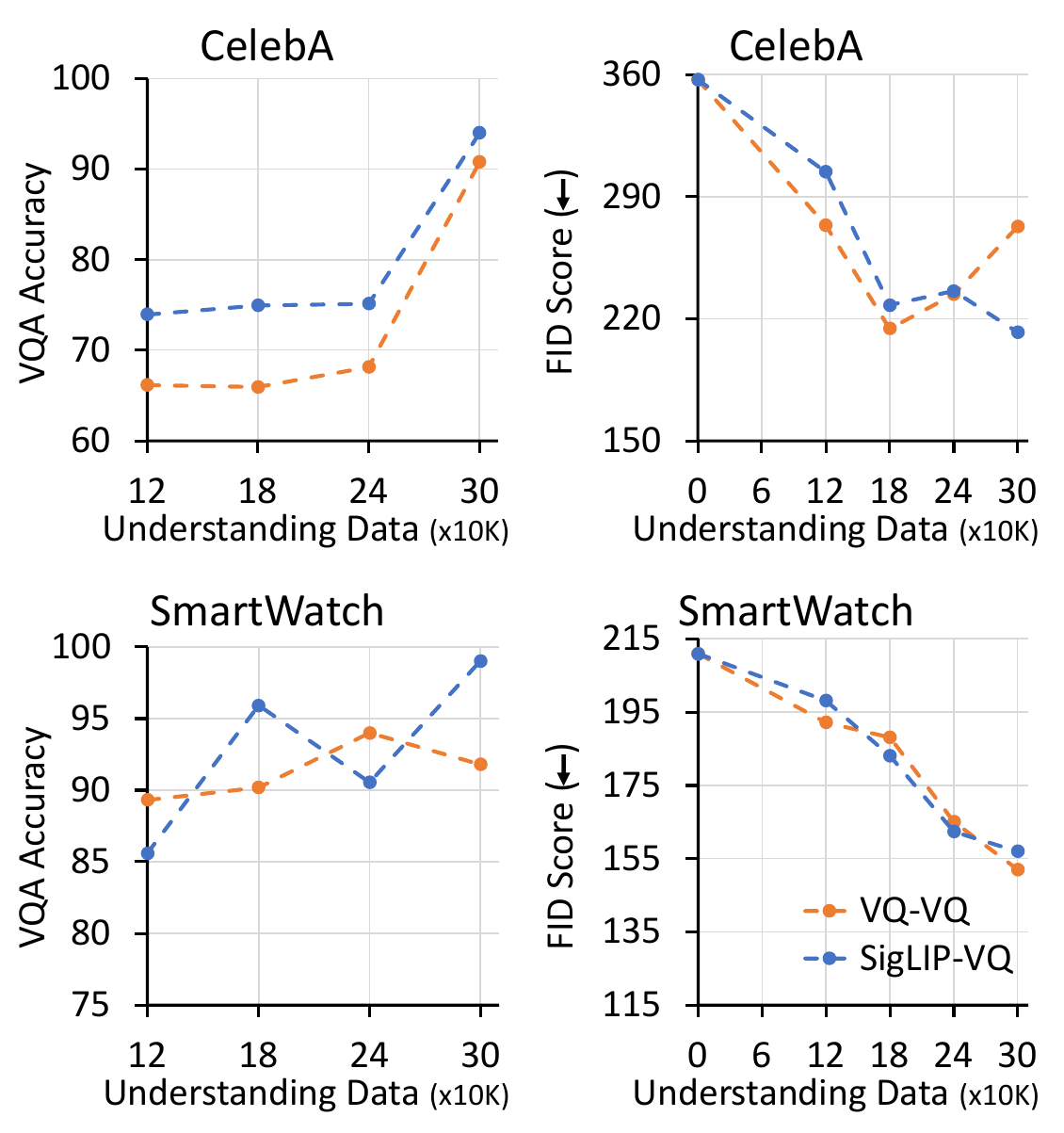}
    }
    \hfill
    \subfigure[Performance of understanding and generation as the amount of generation data increases, with understanding data fixed at 120K.]{
        \label{fig:more_gen}
        \includegraphics[width=0.47\textwidth]{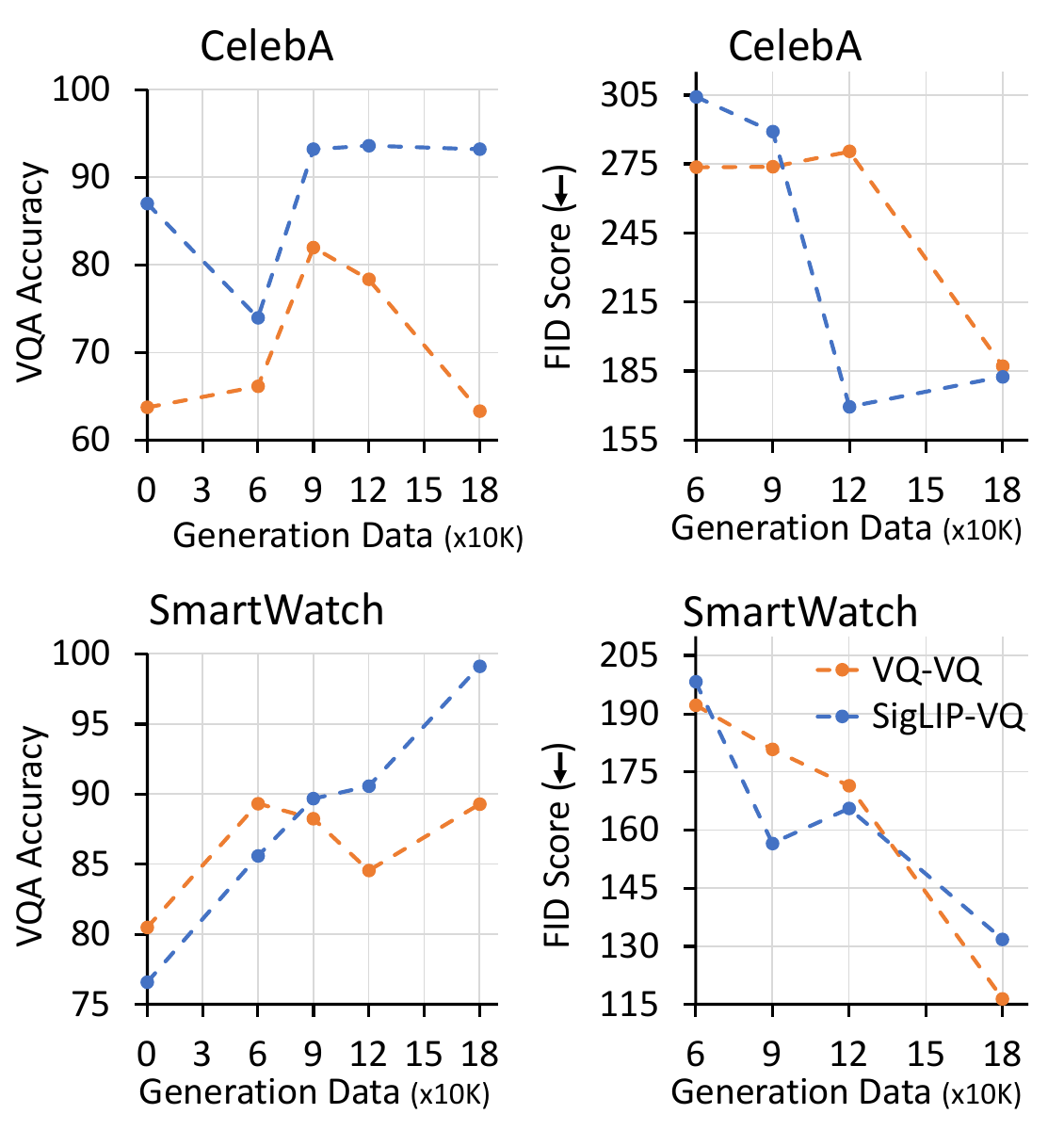}
    }
    \caption{Performance of SigLIP-VQ and VQ-VQ under varying data scales. Adding data from the other task can initially improve performance, but the trend is not always monotonic and can depend on architecture and dataset.}
    \label{fig:eval-2}
\end{figure}

\subsection{Scaling Up with Increased Data}

Building on the previous findings, we investigate how the balance between understanding and generation data affects cross-task generalization. We vary the data mixture in two directions: (1) fixing generation data at 60K while increasing understanding data from 0K to 120K, 180K, 240K, and 300K; and (2) fixing understanding data at 120K while increasing generation data from 0K to 60K, 90K, 120K, and 180K. The results are shown in Figure~\ref{fig:eval-2}.

The results reveal two trends. First, adding data from one task can improve the other task: increasing understanding data can improve generation FID, and increasing generation data can improve VQA accuracy. This supports the existence of cross-task transfer. Second, the improvement is not always monotonic. In several cases, performance improves initially and then saturates or declines as the data mixture becomes imbalanced. A notable exception is SigLIP-VQ on CelebA, where adding generation data can hurt understanding, consistent with the misalignment between its input and output visual spaces. These observations suggest that mixed training is most useful when the two tasks are balanced and the architecture supports transfer between their visual spaces.

These trends clarify why mixed training can appear beneficial in some settings but harmful in others. The early improvement indicates that supervision from one task can provide useful signal for the other, even when no additional target-task data is added. However, once the mixture becomes dominated by one task, optimization may emphasize the dominant objective at the expense of the complementary task. This is not only a matter of sample count: changing the mixture also changes which prediction targets dominate the shared LLM updates. Thus, cross-task transfer is constrained by both data balance and representation alignment: balanced data can expose useful shared structure, while imbalance or misalignment can turn the same mixed-training setup into a source of interference.

\subsection{Knowledge Transfer from Generation to Understanding}

\begin{figure}[!ht]
    \centering
    \subfigure{
        \includegraphics[width=0.47\textwidth]{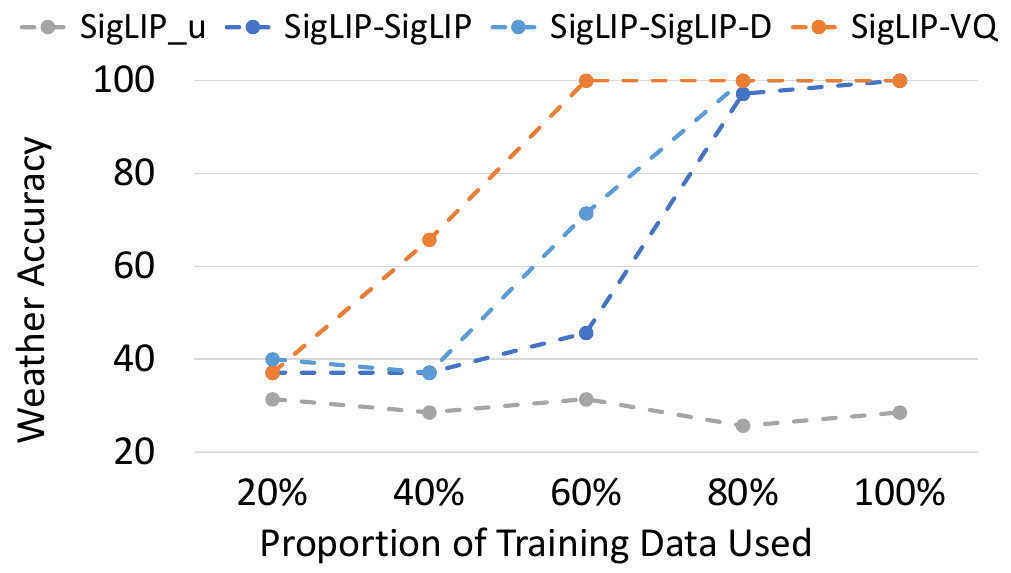}
    }
    \hfill
    \subfigure{
        \includegraphics[width=0.47\textwidth]{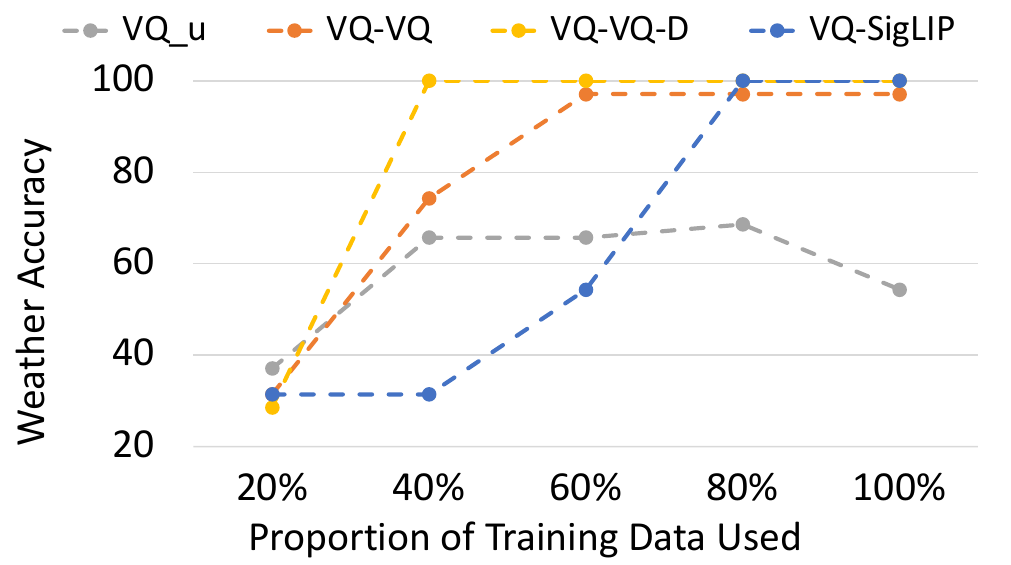}
    }
    \caption{Performance comparison between mixed-training VLMs and understanding-only baselines when weather attributes are underrepresented in understanding data. ``\_u'' denotes understanding-only training. ``-D'' denotes using separate understanding and generation vision adapters.}
    \label{fig:eval-w-biased}
\end{figure}

\begin{figure}[!ht]
    \centering
    \subfigure{
        \includegraphics[width=0.47\textwidth]{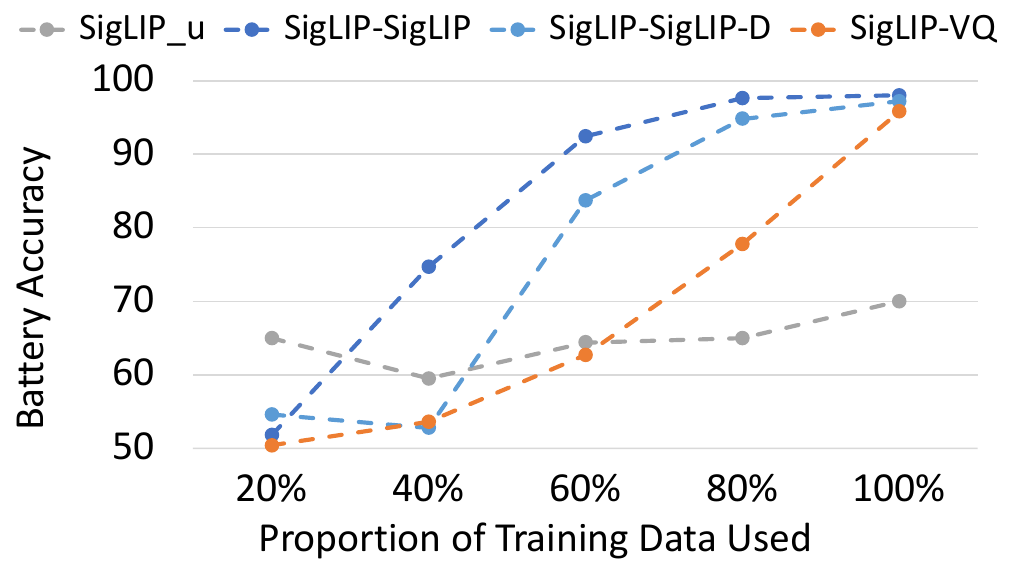}
    }
    \hfill
    \subfigure{
        \includegraphics[width=0.47\textwidth]{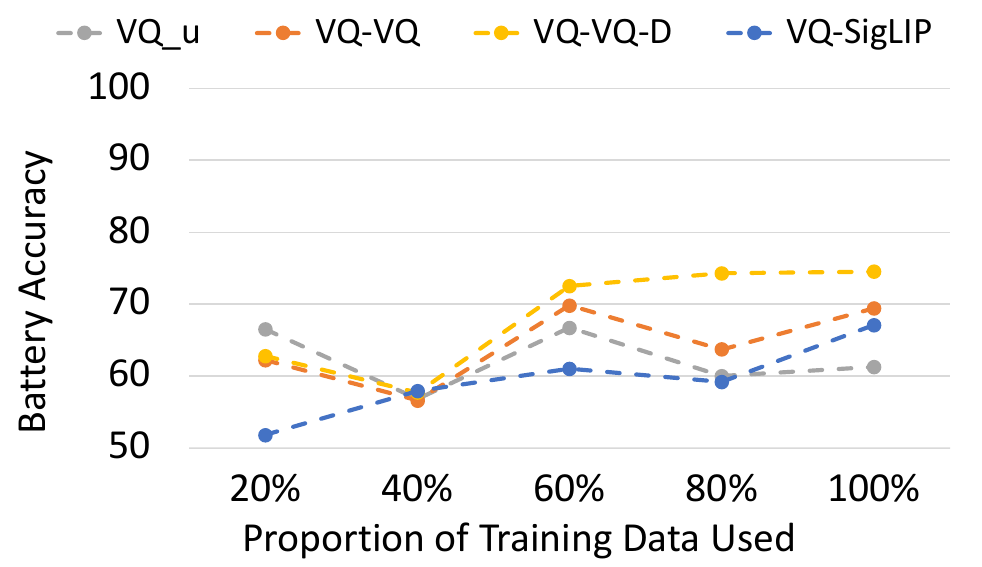}
    }
    \caption{Performance comparison between mixed-training VLMs and understanding-only baselines when battery attributes are underrepresented in understanding data. ``\_u'' denotes understanding-only training. ``-D'' denotes using separate understanding and generation vision adapters.}
    \label{fig:eval-b-biased}
\end{figure}

Based on the finding that generation data can improve understanding in some settings, we next test whether generation supervision can transfer specific visual concepts to understanding. We use the controllability of SmartWatch to reduce the occurrence of one attribute in understanding tasks to a very low level: zero appearances in captioning data and a 0.05 probability of appearance in VQA data, while keeping the generation data unchanged. We apply this manipulation to weather and battery attributes, creating two biased datasets. We then compare understanding-only baselines with mixed-training models. The results are shown in Figures~\ref{fig:eval-w-biased} and~\ref{fig:eval-b-biased}.

As shown in Figure~\ref{fig:eval-w-biased}, when weather is underrepresented in understanding data, both understanding-only models (VQ\_u and SigLIP\_u) struggle to identify weather-related information, while mixed-training models recover the attribute much more effectively. The battery-biased setting in Figure~\ref{fig:eval-b-biased} shows a similar but weaker pattern: SigLIP-based mixed models substantially outperform their understanding-only counterpart, whereas VQ-based models remain more limited. These results suggest that generation supervision can transfer underrepresented visual concepts to understanding, although the strength of this transfer depends on the visual representation.

\begin{figure}[!ht]
    \centering
    \includegraphics[width=\textwidth]{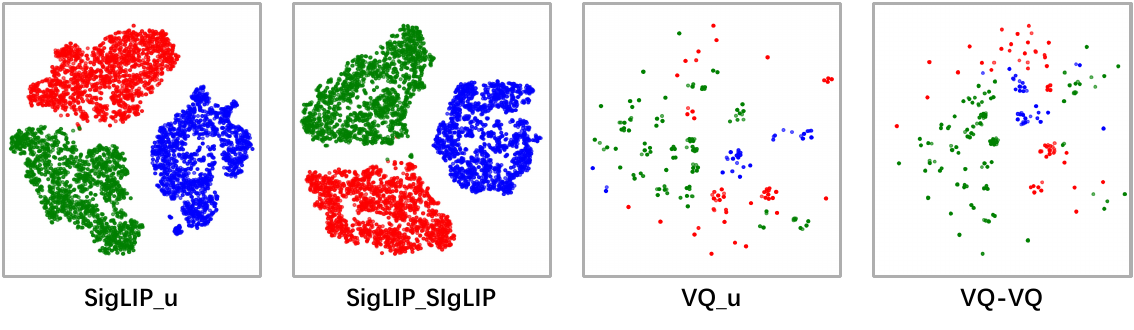}
    \caption{t-SNE visualization of input vision tokens corresponding to the ViT patch representing the weather icon, output by the understanding vision adapter. Samples are colored by weather ground truth, where red, green, and blue refer to sunny, cloudy, and rainy. 5000 samples are shown.}
    \label{fig:tsne}
\end{figure}

To further investigate how this transfer occurs, we analyze the understanding vision tokens output by the adapter in SigLIP\_u, SigLIP-SigLIP, VQ\_u, and VQ-VQ trained on the weather-biased dataset. For each model, we sample 5000 tokens corresponding to the ViT patch that contains the weather icon. Figure~\ref{fig:tsne} visualizes these tokens with t-SNE and colors them by weather label. We further train a linear probe on a 4K/1K train/test split. After 10 epochs, tokens from all four models, including the understanding-only baselines, achieve 100\% linear probing accuracy. This indicates that weather-related information is already present in the input vision tokens even without generation training.

This analysis suggests that the transfer is not primarily caused by generation training making the adapter encode the missing attribute. Instead, the information exists in both mixed-training and understanding-only models, but the understanding-only base LLM does not learn to use it under biased supervision. Mixed generation training appears to provide additional supervision for the relationship between visual tokens and text attributes, allowing the base LLM to reuse relationships learned from the generation side for understanding.

We also test models with separate understanding and generation vision adapters instead of shared adapters. As shown by \textbf{SigLIP-SigLIP-D} and \textbf{VQ-VQ-D} in Figures~\ref{fig:eval-w-biased} and~\ref{fig:eval-b-biased}, transfer can still occur without parameter sharing, which supports that the base LLM, rather than adapter sharing alone, plays an important role in connecting the two task spaces.

\subsection{Applications on Real-Case VLMs}

To examine whether the controlled observations have signs of transfer in a real-case VLM, we conduct experiments based on LLaVA-V1.5-7B~\cite{liu2023visual}. We extend LLaVA by adding a generation vision adapter and a generation vision head, enabling it to generate image CLIP embeddings~\citep{tong2024metamorph} or VQ-VAE token IDs~\cite{chen2025janus}.
For VQ-VAE we use \texttt{vq\_ds16\_t2i}, as in the previous experiments, but use a 2-layer MLP generation adapter instead of a single linear layer because the real-image data is more complex.
For the image generation data, we reverse the image-caption pairs from ShareGPT4V~\citep{chen2023sharegpt4v}. Specifically, we sample 350K image-caption pairs from the 1.2M ShareGPT4V-PT dataset, using the image captions as generation instructions and the corresponding images as generation targets.

Following the two-stage training process of the original LLaVA framework, we proceed as follows:
In the first stage, we freeze the base LLM and vision encoder while updating the understanding vision adapter, generation vision adapter, and generation head. To this end, we augment the original 558K image understanding dataset with an additional 150K image generation samples. 
In the second stage, we unfreeze the vision encoder and update all other parameters. Here, we further augment the original 665K instruction-tuning dataset with 200K image generation samples.
We follow the default LLaVA-1.5 hyperparameters and do not use LoRA in this real-case experiment. Training uses 8 Nvidia H100 80GB GPUs; the pre-training stage finishes within 3 hours and the instruction-tuning stage within 6 hours.

\begin{table}[!t]
    \centering
    \caption{Performance comparison between original LLaVA-1.5-7B (understanding-only) and two mixed-training variants with 350K additional image generation samples (150K in pre-training and 200K in instruction tuning). Results for original LLaVA-1.5-7B are from the official report of LMMs-Eval~\cite{lilmms}.}
    \label{tab:llava}
    \footnotesize
    \begin{tabular}{l c c c c c c c c}
        \toprule
        Model & MME & MMBench\_EN & POPE & VizWiz & MMVet & GQA & MMStar& TextVQA \\
        \midrule
        LLaVA & \textbf{1510.7} & 64.3 & 85.9 & 54.4 & 30.6 & 62.0 &33.3 & 45.8 \\
        CLIP-CLIP & 1506.4 & 65.0 & 86.5 & 55.4 & \textbf{34.4} & 62.0 &36.2 & 47.0 \\
        CLIP-VQ & 1476.6 & \textbf{66.1} & \textbf{86.6} & \textbf{58.1} & 31.2 & \textbf{62.1} & \textbf{36.3} &\textbf{ 47.1} \\
        \bottomrule
    \end{tabular}
\end{table}

We evaluate the MLLMs on eight popular independent MLLM benchmarks~\citep{hudson2019gqa,liu2025mmbench,fu2023mme,chen2024we,yu2023mm,li2023evaluating,gurari2018vizwiz,singh2019towards}. 
As shown in Table~\ref{tab:llava}, adding generation supervision to LLaVA improves most understanding benchmarks. CLIP-CLIP improves six out of eight metrics, with a particularly large gain on MMVet, while CLIP-VQ also improves six out of eight metrics and achieves the best scores on MMBench, POPE, VizWiz, GQA, MMStar, and TextVQA. The two mixed variants use the same additional generation data and the same two-stage training schedule, but their gains appear on different benchmarks. This pattern suggests that generation targets provide different auxiliary signals for understanding: predicting CLIP embeddings emphasizes global semantic image-text alignment, whereas predicting VQ tokens introduces a more fine-grained visual reconstruction target that may better support visually grounded or text-centric benchmarks. Overall, this experiment shows that generation supervision can improve understanding in a practical VLM, and that the form of the generated visual representation affects where the gains appear.

\section{Conclusion}
This work presents a controlled empirical study of cross-task generalization between understanding and generation in unified VLMs. We find that mixed training can improve both tasks over task-specific training, but the benefit depends on architecture, data balance, and especially the relation between vision input and output spaces. Aligned spaces support stronger transfer, whereas reversible input-space distortions weaken the benefit and may cause interference. Generation supervision can also recover underrepresented visual concepts for understanding, with adapter analyses suggesting that the base language model plays an important role in this transfer. A real-case LLaVA experiment provides further evidence that generation training can benefit visual understanding. Overall, these results help explain inconsistent findings in prior unified systems: compatible visual semantics and balanced supervision favor transfer, whereas poorly aligned spaces or imbalanced training can cause interference. Our study is limited to SigLIP and VQ-VAE visual spaces; diffusion-based architectures and broader real-world evaluations remain important directions for future work.

\bibliography{ref}

\end{document}